\pdfoutput=1

\documentclass[11pt]{article}

\usepackage[]{acl}

\usepackage{times}
\usepackage{latexsym}

\usepackage[T1]{fontenc}

\usepackage[utf8]{inputenc}

\usepackage{microtype}

\usepackage{inconsolata}
\usepackage{booktabs}
\usepackage{arydshln}
\usepackage{multirow}
\usepackage{graphicx}
\usepackage{mdframed}

\newcommand{\method}{\textsc{MT-Patcher}\ }
\newcommand{\methodend}{\textsc{MT-Patcher}}

\usepackage{CJK}
\usepackage{xcolor}
\usepackage{ulem}

\newcommand\blfootnote[1]{%
\begingroup
\renewcommand\thefootnote{}\footnote{#1}%
\addtocounter{footnote}{-1}%
\endgroup
}

\title{\textsc{MT-Patcher}: Selective and Extendable Knowledge Distillation from Large Language Models for Machine Translation}

\author{
    Jiahuan Li$^{\clubsuit*}$, Shanbo Cheng$^\spadesuit\dagger$, Shujian Huang$^\clubsuit\dagger$ \and Jiajun Chen$^\clubsuit$ \\
    $^\clubsuit$ National Key Laboratory for Novel Software Technology, Nanjing University, China \\
    $^\spadesuit$ ByteDance Research \\
    \texttt{lijh@smail.nju.edu.cn, \{huangsj,chenjj\}@nju.edu.cn} \\
    \texttt{chengshanbo@bytedance.com}
}

\begin{document}
\maketitle
\blfootnote{$*$ Work done while Jiahuan Li's internship at ByteDance Research.}

\blfootnote{$\dagger$ Corresponding authors.}

\begin{abstract}
Large Language Models (LLM) have demonstrated their strong ability in the field of machine translation (MT), yet they suffer from high computational cost and latency. Therefore, transferring translation knowledge from giant LLMs to medium-sized machine translation models is a promising research direction.  However, traditional knowledge distillation methods do not take the capability of student and teacher models into consideration, therefore repeatedly teaching student models on the knowledge they have learned, and failing to extend to novel contexts and knowledge. In this paper, we propose a framework called \methodend, which transfers knowledge from LLMs to existing MT models in a \textit{selective, comprehensive and proactive} manner. Considering the current translation ability of student MT models, we only identify and correct their translation errors, instead of distilling the whole translation from the teacher. Leveraging the strong language abilities of LLMs, we instruct LLM teachers to synthesize diverse contexts and anticipate more potential errors for the student. Experiment results on translating both specific language phenomena and general MT benchmarks demonstrate that finetuning the student MT model on about 10\% examples can achieve comparable results to the traditional knowledge distillation method, and synthesized potential errors and diverse contexts further improve translation performances on unseen contexts and words. 
\end{abstract}
\section{Introduction}

Large Language Models (LLM) have shown their impressive capabilities across almost all natural language tasks~\citep{gpt3,zhao2023survey}. However, their ability strongly correlates with the model size. In the field of machine translation, competitive results can only be evidenced on larger LLMs, while medium-sized LLMs like Alpaca~\citep{alpaca} and ParroT~\citep{jiao2023parrot} still lag behind supervised NMT systems by a large margin~\citep{jiao2023parrot,zhu2023multilingual}. How to efficiently transfer knowledge from larger LLMs to existing MT models that are affordable to deploy, is an important research direction.

The most common method for knowledge transferring is knowledge distillation (KD)~\citep{hinton2015distilling,kim-rush-2016-sequence}, where given an unlabeled corpus, a student model is trained to mimic the output of a teacher model on the corpus. Although KD is a well-studied technique and has proven effective in many previous works~\citep{kim-rush-2016-sequence,wang-etal-2021-selective,kd_nat}, we argue that when transferring knowledge from giant LLMs to existing MT models, the traditional KD method does not take the capability of the student and teacher model into consideration, therefore leaving much room for improvement in terms of both efficiency and effectiveness.

Firstly, in contrast to student models in previous works~\citep{kim-rush-2016-sequence,wang-etal-2021-selective,kd_nat} that are randomly initialized, recent student MT models~\citep{hsieh-etal-2023-distilling,fu2023specializing} already exhibit a reasonable level of language proficiency, i.e., they can already accurately translate most examples in the unlabeled corpus. This renders the fine-tuning of student models on \textit{all} teacher outputs both redundant and inefficient.

Secondly, the efficacy of KD is significantly constrained by the coverage of the monolingual corpus, which impedes their performance when translating words in novel contexts or words unseen in the monolingual corpus. However, modern LLMs grasp strong translation and language knowledge, as well as the ability to follow human instructions. This enables the development of more efficient and effective strategies for addressing these problems.

In this paper, we introduce \methodend, a novel framework designed for the knowledge transfer from LLMs to existing MT models in a \textit{selective, comprehensive, and proactive} manner. The design philosophy of \method is inspired by effective teaching strategies observed in real-world scenarios. Rather than subjecting students to endless drills, an effective teacher would first assess the student's current abilities, then design practice to reinforce areas of weakness and extend learning to new situations~\citep{lee1979homework,more_than_minutes}. Leveraging the strong language capabilities of LLMs, our method seeks to emulate these pedagogical strategies. Specifically, we gather instructional data from GPT-4, which demonstrates how to identify and correct errors in student model translations, anticipate additional potential errors that the student models may commit, and synthesize diverse contexts for relevant translation knowledge that aids the student model in rectifying these errors. We subsequently fine-tune an existing proficient LLM on these data to transform it into an \method model.

We conduct experiments on translating specific language phenomena (chemistry materials and Chinese idioms) and on general machine translation benchmarks (WMT22 Chinese $\to$ English, English $\to$ German and English $\to$ Japanese). Experimental results show that finetuning the student model on only 10\% examples selected by \method is equivalent to finetuning on all examples as in KD, and enlarging the finetuning corpus via the context synthesis and proactive error prediction technique further improves the translation performance.

\section{Background}

\paragraph{Large Language Model for Machine Translation} 
Numerous studies have attempted to leverage LLMs for machine translation. Initial efforts~\citep{lin-etal-2022-shot,vilar2022prompting,agrawal-etal-2023-context,zhu2023multilingual,hendy2023good,jiao2023chatgpt} centered on in-context learning, which utilizes several translation examples to guide the translation behavior of LLMs. Subsequent research~\citep{jiao2023parrot,li2023eliciting} shifted the focus to fine-tuning LLMs on existing parallel corpora to more effectively harness their translation capabilities. However, the translation performance of LLMs has not been as remarkable as their performance in other NLP tasks. Only state-of-the-art LLMs such as GPT-3 and GPT-4, which boast more than 100 billion parameters, can rival the performance of commercial translation systems~\citep{hendy2023good,jiao2023chatgpt}. Meanwhile, other medium-sized LLMs significantly trail behind supervised MT models~\citep{zhu2023multilingual,li2023eliciting,jiao2023parrot}. \citet{li2023eliciting} suggest that the primary barrier to enhancing LLMs' performance is the lack of translation knowledge. Given that larger LLMs inherently possess more knowledge due to the scaling law~\citep{kaplan2020scaling}, our work concentrates on transferring knowledge from these models to existing MT models.
 
\paragraph{Knowledge Distillation for Neural Machine Translation}
Knowledge distillation (KD), which improves smaller \textit{student} models by learning on larger \textit{teacher} models' output, is widely used in machine translation. Two common KD methods are LogitKD~\citep{hinton2015distilling,tan2018multilingual}, which optimizes the student model to match the teacher model's predicted distribution, and Sequence KD (SeqKD)~\citep{kim-rush-2016-sequence,wang-etal-2021-selective,gu2018non,zhou2019understanding}, where the student learns from the teacher-generated pseudo target sequence. As LogitKD requires access to the teacher's logits, it is impractical for distilling from proprietary LLMs. Therefore, we base our method on SeqKD, where student refers the smaller MT model we would like to improve, and teacher refers to larger LLMs which possess more translation knowledge than student. 

Selective KD has been proposed by \citet{wang-etal-2021-selective} and \citet{kd_nat}, but they all rely on comparing student models' outputs to oracle references. Unlike these works, our method instructs the LLM to identify student translation errors directly.

\paragraph{Large Language Model for Synthesizing Datasets} With the growing generative capabilities of Large Language Models (LLMs), many works attempt to harness them for corpora generation. The generated corpora can serve as demonstrations for few-shot prompting~\citep{sahu-etal-2022-data}, fine-tuning corpora for existing models~\citep{yoo-etal-2021-gpt3mix-leveraging}, or seed corpora for human refinement~\citep{yuan2021synthbio}. Studies such as \citet{chung-etal-2023-increasing,yu2023large} also explore ways to balance diversity, accuracy, and bias reduction in LLM-based dataset synthesis. However, these approaches often generate datasets from scratch, ignoring the capabilities of the models being optimized, resulting in less efficiency compared to our method.

\begin{figure*}
    \centering
    \includegraphics[width=0.8\linewidth]{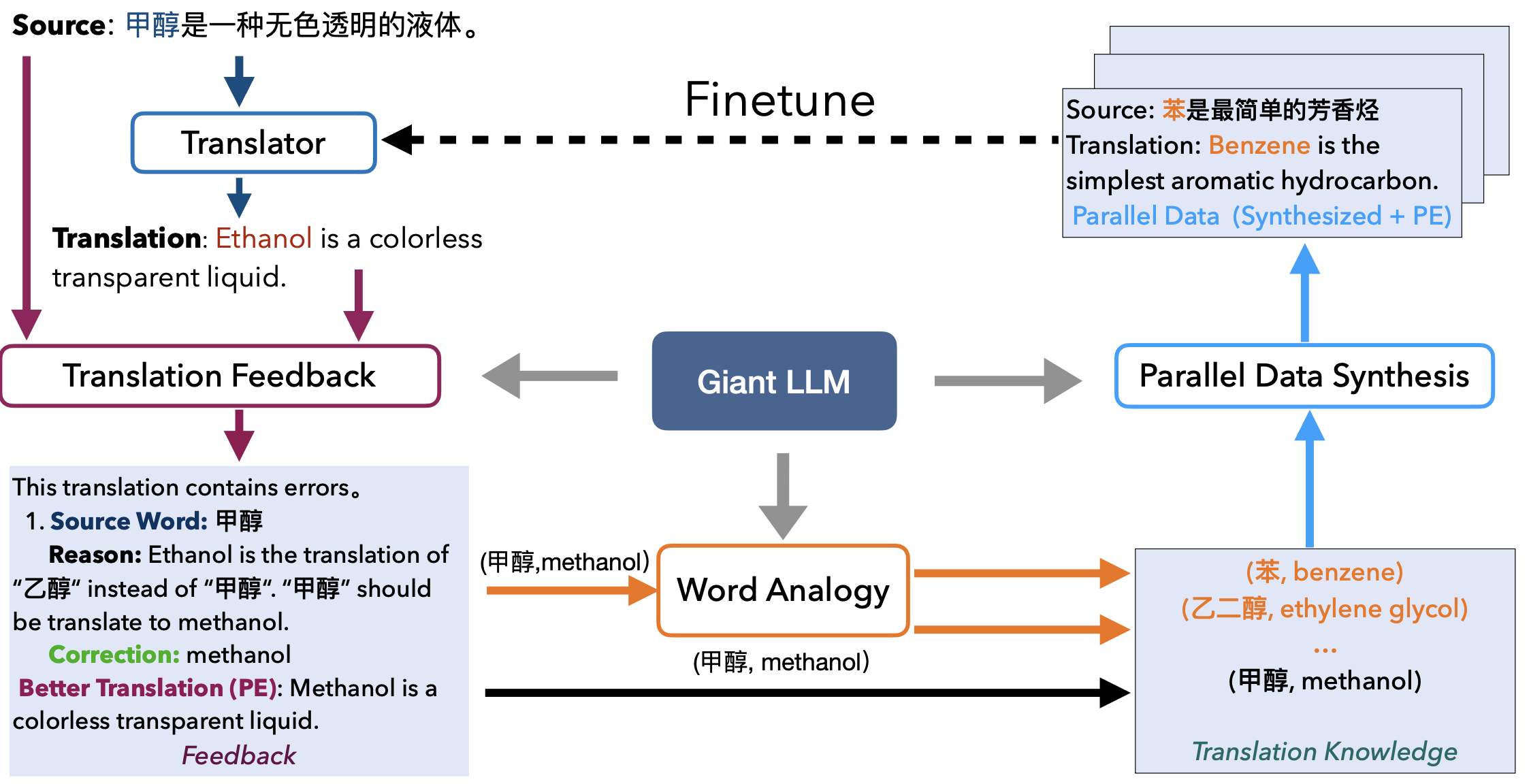}
    \caption{The illustration of \method framework. The correct translation for the source sentence should be `Methanol is a colorless transparent liquid.'.}
    \label{fig:arch}
\end{figure*}

\section{Methodology}
In this section, we present \method, a framework that distills knowledge from LLMs to existing MT systems more efficiently and effectively. The process of \method undergoes two stages:

\begin{itemize}
    \item \textbf{Knowledge Selection}: In this stage, the LLM acts as the \textit{feedbacker}, which provides natural language feedback to translations of student models. Based on the 
 feedback, we select source sentences with identified errors, which indicate knowledge deficiency of the student models, to the next stage.

    \item \textbf{Knowledge Extension}: In this stage, the LLM acts as the \textit{parallel data synthesizer} and \textit{word analoger}, which help the student model learn words it makes mistakes on by extending to more diverse contexts and similar words.
\end{itemize}

Figure~\ref{fig:arch} illustrates how \method works.

\subsection{Knowledge Selection via \textit{Feedbacker}}



When transferring knowledge from LLMs to existing MT models, traditional SeqKD would finetune the student model on \textit{all} teacher's output, ignoring the fact that the student model can already translate most of the examples well.  Furthermore, several recent studies have unveiled emergent abilities in LLMs, such as \textit{Self-Refinement}~\citep{selfrefine} and \textit{Self-Debug}~\citep{selfdebug}, suggesting that iterative refinement of an initial draft may be a more effective strategy to tap into the knowledge reserves of LLMs.

To improve the efficiency of SeqKD and better elicit LLMs' knowledge, we propose to finetune LLMs to be a \textit{feedbacker}, which produces natural language feedback of the student models' translation instead of directly generating its own translations. 
Formally, given a source sentence $X$ and its corresponding translation $Y$, the goal of the feedbacker is to generate a comprehensive assessment $f$. This assessment comprises tuples of $(c,\{(s_i,e_i,t_i)\}_{i=1}^N,p)$, where $c$ describes whether $Y$ contains translation errors, $s_i, e_i, t_i$ corresponds to the source span, explanation and correction of the $i$-th identified error, respectively, and $p$ is the final post-edited translation that incorporates all error corrections.

\subsection{Knowledge Extension via \textit{Parallel Data Synthesizer and Word Analoger}}
Another limitation of SeqKD is that the knowledge it can transfer is strictly limited to the given monolingual corpus. This limitation can hinder its generalizability in two key ways. Firstly, the correct translation of mistranslated words or phrases can only be learned within the contexts present in the given monolingual corpus, potentially limiting its applicability to broader contexts. Secondly, SeqKD also lacks the capacity for knowledge extrapolation, which prevents it from transferring knowledge that does not occur in the monolingual corpus.

Inspired by the principle of knowledge extension when designing good practice in the educational process~\citep{lee1979homework,more_than_minutes}, we transform LLMs into two modules to mitigate above two problems, respectively: \textit{parallel data synthesizer} and \textit{word analoger}.

\paragraph{Parallel Data Synthesizer}
The goal of the parallel data synthesizer is to synthesize parallel sentences $(X',Y')$ that contain a specific pair of phrases $(s,c)$ where the student model makes mistakes in the context $(X,Y)$, in order to generalize the current translation knowledge to more contexts. Ideally, the synthesized parallel sentences should be semantically diverse yet still similar to the original context in other aspects. However, in the preliminary experiments, we find that even for powerful LLMs like GPT-4, when conditioning them on the original context $(X,Y)$, the generated parallel data lacks diversity and mostly resembles $(X,Y)$.

To tackle this problem, we introduce another module called sentence analyzer, which first extracts the information of \textit{domain, topic} and \textit{style} of the original context. We then instruct the LLMs to synthesize parallel sentences with the same attributes as well as containing the phrase pair $(s,c)$. This process can be seen as an information bottleneck where we squeeze the semantic information yet keep other attributes.

\paragraph{Word Analoger} We further introduce the word analoger to proactively predict potential errors the student model may commit. For example, if the student MT model incorrectly translates the term \textit{methanol}, an educated guess is that it may struggle with translating words within the domain of chemistry, such as \textit{benzene} and \textit{ethanol}. By anticipating these potential errors, we can enhance the student model's translation capability for words not present in the monolingual corpus. 

Practically, given a source sentence $X$ and a word $s$ that the student MT model mistranslates, the word analoger aims to associate more words from two perspectives: (1) category, i.e., words belonging to the same category as $s$, and (2) semantic, i.e., words that frequently co-occur with $s$. We also require that the generated words should be rare and challenging in the prompt, ensuring that the student model will struggle to translate them accurately.

\subsection{Implementation of \method}
Theoretically, state-of-the-art LLMs like GPT-4 can already serve as an \method to transfer its knowledge to MT models. However, in practice, because we do not have unlimited access to GPT-4, we instead collect the demonstration data from GPT-4. Specifically, given a student model, we first use it to generate its translation on 20,000 monolingual sentences randomly selected from the monolingual corpus. We then leverage GPT-4 to execute the pipeline of \method including (1) giving feedback $f$ given the source sentence and student's translation $(X, Y)$, (2) analyzing the domain, topic and style $(d,t,st)$ of the source sentence $X$ (3) making analogies $(WA_x,WA_y)$ given the source sentence $X$ and a word $s$ in $X$ (4) synthesizing parallel sentences containing error source words $s$ and their corrections $c$ with the same domain, topic and style attribute $(d,t,st)$. Finally, we finetune the teacher LLM on these data to transform it to an \methodend. All prompts we use for building \method can be found in Appendix~\ref{sec:method_prompt}.
\section{Experiments}


We evaluate our method on Chinese $\to$ English and English $\to$ German translation.

\begin{table*}[t]
\centering
\small
\begin{tabular}{@{}lcccccccc@{}}
\toprule
\multicolumn{1}{c}{\multirow{2}{*}{\textbf{System}}} & \multicolumn{4}{c}{\textbf{Chinese} $\to$ \textbf{English}}          & \multicolumn{4}{c}{\textbf{English} $\to$ \textbf{German}}         \\
\multicolumn{1}{c}{}                        & \multicolumn{4}{c}{\textit{Teacher Model: Baichuan2 13B}} & \multicolumn{4}{c}{\textit{Teacher Model: Llama2 13B}}  \\ \hline
                                            & |$\mathcal{D}_f$|   & COMET & BLEURT  & BLEU & |$\mathcal{D}_f$| & COMET & BLEURT & BLEU \\ \hline
Teacher                            & -    &  80.5 & 67.8 &  23.9 & -    & 81.4 & 72.9  & 26.0\\ \hline

\multicolumn{9}{c}{\textit{Student Model: ParroT-7B}}                                       \\ 
Student                            & -    &  75.4 & 60.6 & 18.1 &  -     & 80.5 & 69.0 & 23.9\\
SeqKD-Equal                           & 119k & 76.0 & 61.4 & 21.9 & 107k  & 80.3 & 70.8 & 24.1 \\ 
SeqKD-Full                            & 1M   &  76.5 & 61.7 & 22.2 & 1M   & 80.9 & 71.4 & 24.6 \\
\hdashline
\textsc{MT-Patcher}                         &      &      &      &  &      &      &      &  \\
\multicolumn{1}{r}{+ PE}           & 119k & 76.7 & 61.8 & 22.4 & 107k & 80.9 & 71.6 & 24.9 \\
\multicolumn{1}{r}{+ PE + PDS}      & 595k & 77.4 & 62.6 & 23.0 & 535k & 81.3 & 72.0 & 25.5 \\
\multicolumn{1}{r}{+ PE + PDS + WA} &  1.07M  & \textbf{78.2} & \textbf{63.5} & \textbf{23.8} & 963k   & \textbf{81.8} & \textbf{72.6} & \textbf{26.2}\\ \midrule

\multicolumn{9}{c}{\textit{Student Model: NLLB 3.3B}}                                       \\ 
Student                            & -    &  76.8 & 63.9 & 20.8 & -    & 86.1 & 76.3  & 34.3 \\
SeqKD-Equal                           & 104k &  79.1 & 66.3 & 25.0 & 124k  & 85.2 & 74.7 & 32.0\\ 
SeqKD-Full                            & 1M   &  79.5 & 66.9 & \textbf{25.5} & 1M   & 84.8 & 74.1 & 31.2 \\
\hdashline
\textsc{MT-Patcher}                         &      &      &      &  &      &      &      &  \\
\multicolumn{1}{r}{+ PE}           & 104k &  79.4 & 67.0 & 24.2 &  87k & 86.2 & 76.5 & 34.5\\
\multicolumn{1}{r}{+ PE + PDS}      & 520k &  79.9 & 67.4 & 24.8 & 435k & 86.5 & 77.0 & 34.9\\
\multicolumn{1}{r}{+ PE + PDS + WA} & 936k   &  \textbf{80.3} & \textbf{68.1} & 25.4 & 783k    & \textbf{87.2} & \textbf{77.5} & \textbf{35.6}\\ 
\bottomrule
\end{tabular}
\caption{Translation performance of the proposed method and other baselines on the WMT22 Chinese$\to$English and English$\to$German test sets. $|D_f|$ denotes the number of examples used to finetune the student model. SeqKD-Full refers to the student model finetunes on the full 1M pseudo parallel sentences, while SeqKD-Equal finetunes on random subsets of the teacher's translations with equal size to that of \methodend.}
\label{tab:main_results}
\end{table*}

\subsection{Experimental Settings}
\paragraph{Student Translation Model}
For student translation models, we consider NLLB-200 3.3B~\citep{nllbteam2022language}, a multilingual translation model pre-trained on 200 languages. Having been trained on massive parallel data, it can already translate reasonably well but falls short of language knowledge compared to LLMs, making it an ideal knowledge recipient for our experiment. 

Due to the increasing interest in adopting LLMs for MT, we also consider ParroT~\citep{jiao2023parrot}, an LLM-based MT model finetuned on WMT validation sets from LLaMA-7B~\citep{touvron2023llama}. 

\paragraph{Backbone LLM for \method}
The backbone LLMs for building \method in this paper are LLaMA2-13B~\citep{touvron2023llama} and Baichuan-2-13B~\citep{baichuan2}. LLaMA2-13B is an English LLM and used to build \method for English-German translation models. Baichuan-2-13B is trained on a mix of both Chinese and English corpus and demonstrates much stronger abilities in Chinese compared to LLaMA2. Therefore, we adopt it for building \method for Chinese-English translation models. For each language pair considered, we fully finetune the corresponding LLM on the collected data for 3 epochs. See Appendix~\ref{sec:hyper_param} for more implementation details.




\paragraph{Competitors}
We compare the translation performance of the following methods:

\begin{itemize}
    \item \textbf{Student} is the translation model to be patched. In this paper, it refers to NLLB 3.3B or ParroT.
    \item \textbf{Teacher} is the model that is achieved by finetuning the larger LLM to perform translation directly. For a fair comparison, we finetune the LLM on GPT-4's translation on the monolingual sentences.
    \item \textbf{SeqKD} are models achieved by finetuning the Student model on the Teacher's translations. 
    \item \textbf{\method~(PE)} is the variant of \method, finetuning the Student model on the post-editing results in feedback.
    \item \textbf{\method~(PE + PDS)} is the variant of \method which finetunes the Student model on the post-editing results as well as additional synthesized parallel sentences generated by parallel data synthesizer containing (error, correction) pairs. Unless other stated, we set the number of pseudo-parallel sentences to be 4 in this paper.
    \item \textbf{\method~(PE + PDS + WA)} is the variant of \method which finetunes the Student model on the post-editing results and parallel sentences generated by parallel data synthesizer containing (error, correction) pairs and additional word pairs from word analoger. We generate 2 analogous words for each category and 1 context for each word.
\end{itemize}

\begin{table*}[t]
\centering
\small
\begin{tabular}{@{}lcccccccc@{}}
\toprule

\multicolumn{1}{c}{}          & \multicolumn{4}{c}{\textbf{Chemistry Materials}}            & \multicolumn{4}{c}{\textbf{Chinese Idioms}}          \\ \midrule
 & \multicolumn{2}{c}{\textit{Unseen Context}} & \multicolumn{2}{c}{\textit{Unseen Word}} & \multicolumn{2}{c}{\textit{Unseen Context}} & \multicolumn{2}{c}{\textit{Unseen Word}} \\
                                    & Accuracy & Rel. Perf. & Accuracy & Rel. Perf.  & Score & Rel. Perf. & Score & Rel. Perf. \\ \hline
Student                             & 6.0        & 22.4\%          & 6.3      & 23.7\%          & 1.20  & 39.8\%       & 1.16  & 37.4\%       \\
Teacher                             & 26.0       & 97.4\%        & 25.8     & 97.4\%        & 2.78  & 92.3\%        & 2.82  & 91.0\%        \\ \hdashline
Feedbacker                          & 26.7     & 100\%      & 26.5     & 100\%      & 3.01  & 100\%      & 3.10  & 100\%     \\ \hdashline
SeqKD-Full                          & 15.5     & 58.1\%       & 10.6     & 40.0\%       & 1.65  & 54.8\%       & 1.62  & 52.3\%       \\
\method                          &          &              &          &              &       &              &       &              \\
\multicolumn{1}{r}{+ PE}            & 15.8     & 59.2\%       & 11.0       & 41.5\%       & 1.73   & 57.5\%       & 1.78  & 57.4\%       \\
\multicolumn{1}{r}{+ PE + PDS}      & 21.4     & 80.5\%       & 11.2     & 42.3\%       & 2.04  & 67.8\%       & 1.81  & 58.4\%       \\
\multicolumn{1}{r}{+ PE + PDS + WA} & 21.9     & 82.0\%       & 16.3     & 61.5\%       & 2.10  & 69.8\%       & 2.02  & 65.2\%       \\ \bottomrule
\end{tabular}
\caption{Performance of different models when translating chemistry materials (evaluated in accuracy) and Chinese Idioms (evaluated by scores given by GPT-4). Rel. Perf: the relative performances of models compared to feedbacker, which is the best extent we can elicit knowledge from LLMs in this table.}
\label{tab:idiom}
\end{table*}

\begin{table}[t]
\centering
\small
\begin{tabular}{lccc}
\toprule
                       & BLEU & COMET & BLEURT \\ \midrule
Student         & 15.4 & 85.1  & 58.6   \\
SeqKD                  & 16.3 & 85.7  & 61.6   \\
\method & \textbf{16.8} & \textbf{86.4}  & \textbf{62.2}   \\
\bottomrule
\end{tabular}
\caption{Effectiveness of \method on WMT English $\to$ Japanese translation test sets. The student model is NLLB 3.3B.}
\label{tab:en_ja}
\end{table}

\subsection{Results on General Machine Translation}

Table~\ref{tab:main_results} presents experimental results on general machine translation benchmarks: WMT22 Chinese$\to$English and English$\to$German translation. We randomly select 1,000,000 sentences from RefinedWeb~\citep{refinedweb} and WuDao 2.0~\citep{wudao}, respectively, as English and Chinese monolingual corpus. Performance are evaluated in COMET~\citep{rei-etal-2020-comet}, BLEURT~\citep{sellam-etal-2020-bleurt}~\footnote{The model we used for COMET and BLEURT is wmt22-comet-da and BLEURT-20, respectively.} and sacreBLEU~\citep{post-2018-call}. We can see that:

\paragraph{\method can select more valuable examples.} From Table~\ref{tab:main_results}, we can first see that the performance of \method (PE) is better SeqKD-Equal, and can be comparable to SeqKD-Full. This indicates the proposed method can select more valuable examples and discard useless examples. We also find our method suffers less from catastrophic forgetting compared to SeqKD-Full (See Appendix~\ref{sec:catastrophic_forgetting} for more experimental results). This makes \method an appealing method for real-world applications, considering the cost for finetuning the Student model is growing nowadays.

\paragraph{Parallel data synthesizer and word analoger improve the effectiveness of \methodend.} We can also see that applying the parallel data synthesizer and word analoger to generate more patch data can further improve the translation performance of \methodend, highlighting the benefits of extending coverage of context and knowledge during the process of knowledge transferring.

\paragraph{} It is worth noting that in the English $\to$ German direction, the teacher based on LLaMA-2-13B performs substantially worse than the student~(NLLB 3.3B), which is consistent with previous findings~\citep{li2023eliciting} that it is not trivial to adopt existing LLMs to outperform supervised translation models. As a result, SeqKD from this teacher leads to poor performance. However, based on the same backbone LLM, \method can still improve the performance of the Student model. This can be attributed to the hypothesis that revising an initial draft is a better way to elicit the knowledge of LLMs than direct generation, which we provide a further analysis in Section~\ref{sec:dg}.

\paragraph{\method also works when the teacher is not very strong.} Although we mainly focus on settings where we have strong teachers~(which is why we choose different teacher models for English $\to$ German and Chinese $\to$ English translation), we also experiment with medium resource translation: WMT22 English $\to$ Japanese, using LLaMA2 as the teacher and NLLB 3.3B as the student. We present the results in Table~\ref{tab:en_ja}. We find \method can still outperform SeqKD in this setting.

\subsection{Results on Specific Language Phenomena}

In order to understand how \method can improve the effectiveness of knowledge transfer, we present experiments on the Chinese-to-English translation for two specific language phenomena: \textit{chemistry materials} and \textit{Chinese idioms}. We select them for two reasons: \textbf{(1)} Both belong to long-tailed knowledge that student MT models cannot grasp very well. \textbf{(2)} There are also distinctions between them: chemistry materials represent simple, context-free knowledge, while Chinese idioms represent more abstract and metaphorical knowledge.

Specifically, for each language phenomenon, we first collect a list of 6,000 of them and their corresponding translations from the web. We then split these word pairs into two categories: \textit{Seen} and \textit{Unseen}, and create a monolingual set as well as two test sets based on the split~\footnote{Details of the dataset and data split can be found in Appendix~\ref{sec:dataset_detail}.}: 

\begin{itemize}
    \item \textbf{Monolingual Set}. For each word pair in the \textit{Seen} set, we ask GPT-4 to synthesize one sentence that contains the source word. This set is for SeqKD and \method to leverage.
    \item \textbf{Test Set for Unseen Context.} For each word pair in the \textit{Seen} set, we also ask GPT-4 to synthesize one parallel sentence pair that contains the source and target word in the source and target sentence, respectively. This set is for testing models' generalization ability when source words are seen yet contexts are novel.
    \item \textbf{Test Set for Unseen Word.} We collect the test set for Unseen Word in a similar way as Unseen Context using the word pairs in the \textit{Unseen} set. This set is for testing models' generalization ability to novel words.
\end{itemize}

We take the Baichuan-2-13B as the LLM and NLLB 3.3B as the student model, and present the experimental results in Table~\ref{tab:idiom}. The accuracy of translating chemistry materials represents the percentage of test examples where the correct translation of the source chemistry material is found in the translation. Regarding Chinese idioms, due to the difficulty of providing reference translations of them, we instead ask GPT-4 to assess the translation quality given the source sentence, target sentence and dictionary definition. We report the average score, which ranges from 0 to 5. For ease of comparison, we also report how different models perform relative to the feedbackers, for which we directly take its correction as the translation.

\paragraph{Multiple contexts facilitate generalization on \textit{Unseen Context}.} From Table ~\ref{tab:idiom}, we can see that despite that the Teacher model achieves significantly better performance than the Student model, the SeqKD-Full method can only narrow less than half of the gap. However, by synthesizing more contexts for each error, \method (+PE + PDG) improves the relative performance from 59.2\% to 80.5\%  for chemistry materials, and 57.5\% to 69.8\% for Chinese Idioms, indicating the importance of translation knowledge in multiple contexts in order to generalize to novel contexts better.

\paragraph{Error Anticipation improves performances on \textit{Unseen Word}.} We can also observe that both SeqKD-Full and \method (+PE + PDG) cannot behave well on the \textit{Unseen Word} set, which can be attributed to their inability to extrapolate from the observed errors to unseen errors. By generating analogous words to anticipate more errors, the translation performances on \textit{Unseen Word} are significantly improved, validating the effectiveness of the proposed error anticipation method.

\section{Discussion}

We provide further analysis on how \method works and its applicability to real-world scenarios. All experiments are conducted on the WMT22 Chinese-to-English translation datasets, and the student MT model is NLLB 3.3B.
\begin{figure}
    \centering
    \includegraphics[width=0.8\linewidth]{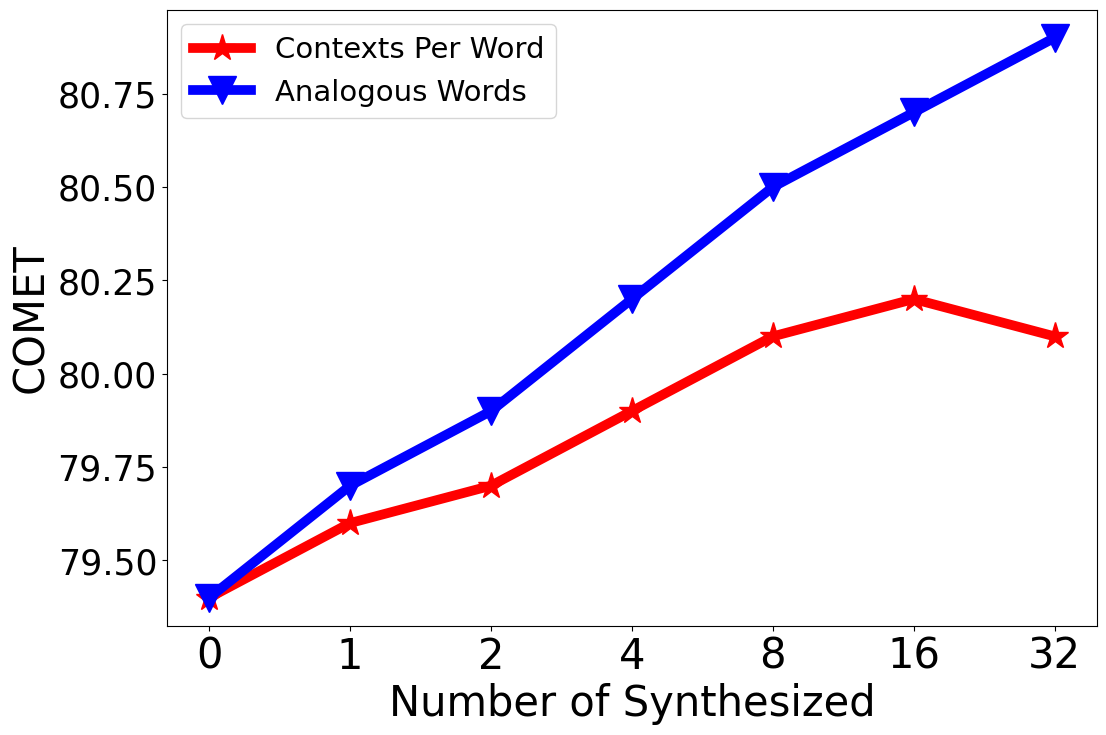}
    \caption{Translation performance as the number of synthesized contexts per word and analogous word grows.}
    \label{fig:scaling}
\end{figure}

\subsection{Impact of the number of synthesized contexts per word and analogous word}

In Figure~\ref{fig:scaling}, we plot how increasing the number of synthesized contexts per word and analogous words affects the translation performance of the student model. Note that we only synthesize one context for each analogous word. We can see increasing both numbers results in improved translation performance.  For synthesized contexts, the gain plateau between 16 to 32 suggests this amount of different contexts is adequate for word or phrase learning. For analogous words, however, we observe the performance grows at a log-linear rate~\footnote{It is worth noting that this does not mean \method can improve the translation performance endlessly, since it cannot generate an unlimited amount of valid analogous words. The performance will eventually plateau, although we have not scaled to the number due to the computational limitation.}. 

\begin{figure}
    \centering
    \includegraphics[width=0.8\linewidth]{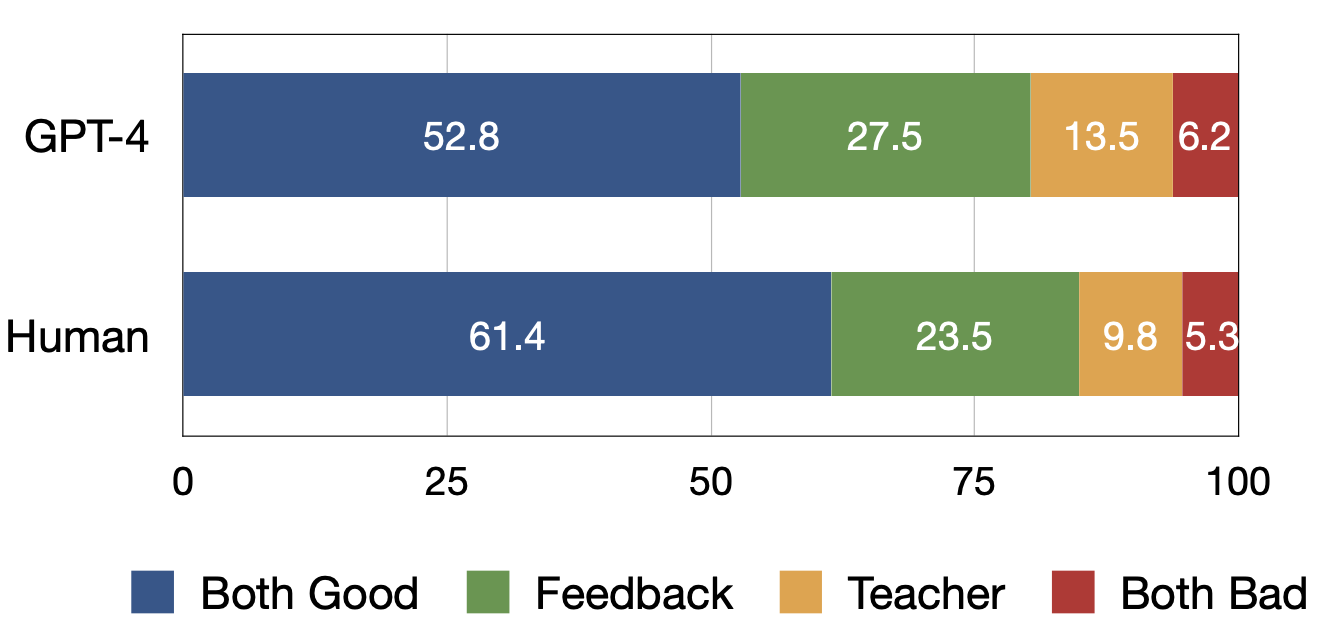}
    \caption{Comparison of translation quality on error words between the Teacher's translation and the feedbacker's correction.}
    \label{fig:dg_vs_feedback}
\end{figure}

\subsection{Does asking for feedback better elicit LLMs' translation knowledge?}
\label{sec:dg}
We conduct a head-to-head comparison between two ways to leverage the teacher LLM: ask the teacher to directly provide translation vs. ask \method to give feedback on the student's translation. Specifically, we randomly select 1000 examples and compare the correction provided by \method to the translation provided by the teacher. The comparison is made by both human and GPT-4. 

The results are shown in Figure~\ref{fig:dg_vs_feedback}. It can be seen that \methodend's corrections are considered by both GPT-4 and human evaluators to be comparable or better than the teacher's translation on more than 80\% examples, demonstrating the benefits of eliciting LLM's knowledge in the form of feedback.

\subsection{The Effectiveness of Iterative Feedback}

\begin{figure}
    \centering
    \includegraphics[width=0.8\linewidth]{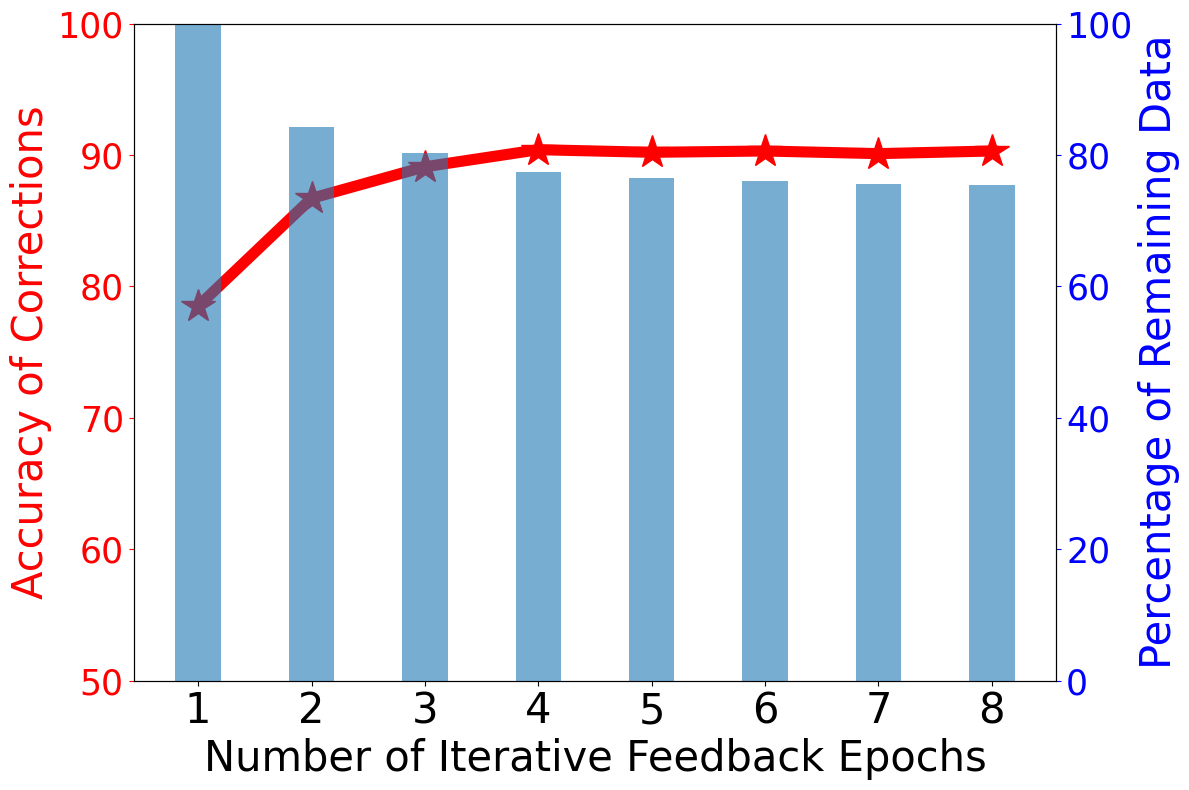}
    \caption{Accuracy of corrections and percentage of remaining data after applying different epochs of iterative feedback.}
    \label{fig:iterative}
\end{figure}

\begin{table}[]
\centering
\small
\begin{tabular}{lccc}
\toprule
 & COMET & BLEURT  & BLEU  \\ \midrule
$k=1$ & 79.4  & 67.0  & 24.2  \\
$k=2$  & 79.8  & 67.5 & 24.7  \\
$k=3$  & 80.0  & \textbf{67.6}  & 24.9 \\
$k=4$  & \textbf{80.1}  & \textbf{67.6} & \textbf{25.1} \\ 
$k=5$  & \textbf{80.1}  & 67.5 & 25.0 \\ 
$k=6$  & 80.0  & \textbf{67.6} & 24.8 \\ 
$k=7$  & 79.8  &  67.4 & 24.9 \\ 
$k=8$  & \textbf{80.1}  & \textbf{67.6} & 24.9 \\ 

\bottomrule
\end{tabular}
\caption{Translation performance of NLLB-3B model finetuned on post-editing data after $k$ epochs of iterative feedback.}
\label{tab:iterative}
\end{table}
In this section, we explore whether the application of iterative feedback on post-edited translations can enhance the final translation quality, thereby yielding a better Student model. While iterative feedback may incur additional computational costs, it allows us to compare feedback across multiple iterations and assess the reliability of error identification and correction from the feedbacker. Intuitively, if an error span identified and rectified in the $i$-th epoch is still deemed problematic in the subsequent epoch, it suggests an inconsistency in the feedbacker's decision-making process. To prevent the introduction of incorrect knowledge during the knowledge transfer process, examples with such inconsistencies are discarded.

We randomly select 2000 instances of \methodend's feedback on NLLB-3B's translation results and apply iterative feedback. We then ask GPT-4 to evaluate the feedback quality after each iterative feedback epoch. The results, depicted in Figure~\ref{fig:iterative}, indicate that iterative feedback can enhance the accuracy of corrections in remaining examples, converging to 90.4\% after 4 epochs at the expense of filtering out approximately 20\% of examples. To understand the quality-quantity trade-off of demonstration data, we further fine-tune the Student NLLB model on post-editing data after each iterative feedback epoch and display the translation performance in Table~\ref{tab:iterative}. Despite a decrease in the amount of fine-tuning data as the epoch increases, the translation performance of the fine-tuned model continues to improve, highlighting the significance of high-quality fine-tuning data.

\begin{table}[]
\small
\begin{tabular}{@{}lcccc@{}}
\toprule
                   & \multicolumn{2}{c}{NLLB} & \multicolumn{2}{c}{ParroT} \\ \midrule
                   & \textsc{ZH}$\to$\textsc{EN}    & \textsc{EN}$\to$\textsc{DE}   & \textsc{ZH}$\to$\textsc{EN}     & \textsc{EN}$\to$\textsc{DE}    \\
Student &  76.8 & 86.1 & 75.4 & 80.5 \\
SeqKD-Full  &   79.5 & 84.8 & 76.5 & 80.9 \\ \hdashline  
NLLB$^\dagger$   &       80.3      &     87.2      &        77.5      &     81.3        \\
ParroT$^\dagger$ &      79.9       &     86.8       &       78.2       &      81.8       \\ \bottomrule
\end{tabular}
\caption{Translation performances when applying \method trained on one student model to another. Performances are evaluated by COMET score. Models with $\dagger$ are \method  (+ PE + PDS + WA)  trained for the corresponding MT model. For reference, we also list the performances of the original student model and SeqKD-Full baselines.}
\label{tab:transfer}
\end{table}

\subsection{Transferability of \method}

The construction of \method is model-dependent; that is given an MT model, LLMs are finetuned on the data from GPT-4 which demonstrates how to execute the \method pipeline on the translation of the corresponding MT model. Considering the cost of data collection and model training, one may question whether \method is transferable, i.e., a patcher model for one MT model can improve the performance of another MT model. We present such results in Table~\ref{tab:transfer}. Although the performance of applying \method to its dedicated MT model is superior, the application of \method trained on another model still significantly surpasses the baseline results, suggesting the potential for a robust \method across various MT models.
\section{Conclusion}
We introduce \methodend, a framework designed to leverage capabilities of LLMs to enhance the efficiency and effectiveness of translation knowledge transfer from LLMs to existing MT models. Our approach involves a pipeline that initially generates feedback on translations produced by MT models, followed by the synthesis of potential errors and diverse contexts to systematically rectify these translation errors. Through experimentation on both general and narrow domain MT benchmarks, we demonstrate that \method effectively improves student MT models' performances compared to SeqKD baselines, and exhibits successful transferability across different models.

In the future, we plan to refine our method from two angles. Firstly, previous works~\citep{freitag-etal-2019-ape,riley-etal-2020-translationese} have identified translationese as a significant issue, and training on pseudo data generated by LLMs can exacerbate this problem. A promising solution could involve retrieving target sentences containing correction words and back-translating them to the source side. Secondly, the feedback's \textit{reason} field contains a wealth of valuable information. We intend to explore more efficient strategies to harness this data.

\section*{Limitations}
Our method focuses on transferring translation knowledge, especially long-tailed lexical knowledge from LLMs to existing MT models, which cannot solve all kinds of translation errors, such as misunderstanding the sentence structure, over/under-translation, etc.

We leverage GPT-4 as evaluators in multiple experiments in this paper. Despite its evaluation has been shown to correlate with human beings well in many previous works, there is still knowledge deficiency in itself and cannot guarantee that the evaluation contains no errors.

\section*{Acknowledgement}
We would like to thank the anonymous reviewers for their insightful comments. Shujian Huang and Shanbo Cheng are the corresponding authors. This work is supported by National Science Foundation of China (No. 62376116, 62176120), the Liaoning Provincial Research Foundation for Basic Research (No. 2022-KF-26-02).

\bibliography{anthology,custom}
\newpage
\appendix
\begin{figure*}
    \centering
    \includegraphics[width=0.8\linewidth]{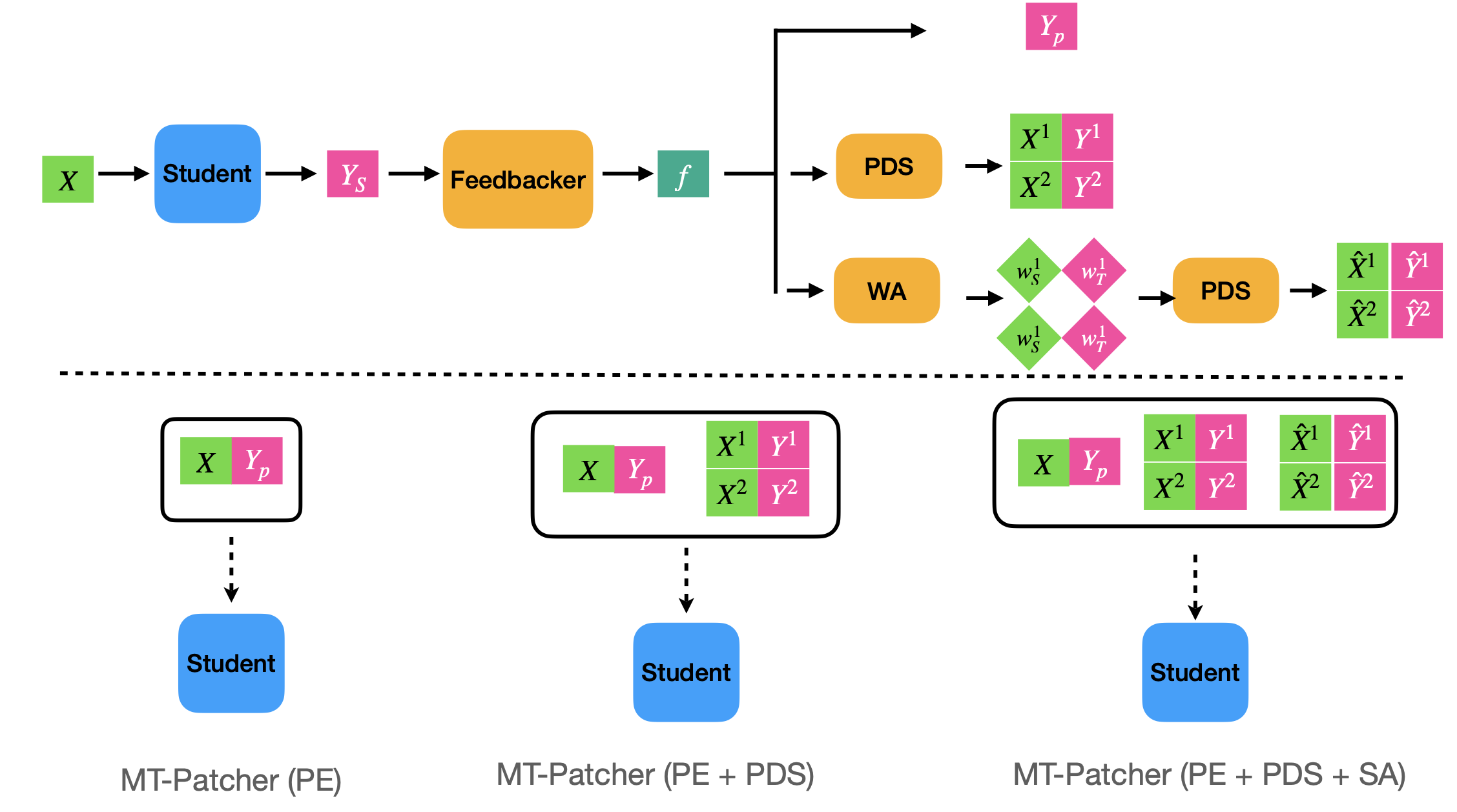}
    \caption{Illustration of variants of \methodend. PDS denotes the parallel data synthesizer, and WA denotes the word analoger.}
    \label{fig:enter-label}
\end{figure*}

\section*{Appendix}

\section{Prompts for \method}
\label{sec:method_prompt}
Table~\ref{prompt:feedback}, \ref{prompt:sentence_analysis}, \ref{prompt:pds}, \ref{prompt:wa} shows the prompt we used for the feedbacker, sentence analysis, parallel data synthesis and word analogy task, respectively.

\begin{table}[]
\small
\centering
\begin{tabular}{lccc}
\toprule
                & \textbf{COMET} & \textbf{BLEURT} & \textbf{BLEU} \\
                \midrule
Student         & 82.4                      & 70.4                       & 26.4                     \\
SeqKD-Full      & 75.9                      & 62.8                       & 22.3                     \\
\method & 81.7                      & 69.5                       & 26.3  \\
\bottomrule
\end{tabular}
\caption{Translation performance on WMT22 German $\to$ English test set. SeqKD-Full and \method are finetuned student models on pseudo Chinese $\to$ English parallel sentences.}
\label{tab:catastrophic_forgetting}
\end{table}

\section{Implementation details}
\label{sec:hyper_param}
We fully finetune LLMs on the collected demonstration data from GPT-4 for 3 epochs. The learning rate is set to $1e-5$, and the batch size is 64. During training, we only compute the next token prediction loss on the response tokens.

\section{\method suffers less from catastrophic forgetting.}
\label{sec:catastrophic_forgetting}

We test the German$\to$English performance of competitors in the Chinese$\to$English setting, including the original student model~(ParroT-7B), SeqKD-Full, and \method (PE). We found SeqKD-Full experiences a significant decrease in performance, while MT-Patcher's performance degradation is much less. This suggests that \method is less prone to catastrophic forgetting, thereby demonstrating its potential for repeated application to a target MT system without detriment to its initial capabilities.

\section{Details of datasets used for chemistry materials and Chinese idioms}
\label{sec:dataset_detail}
For chemistry materials, the data is extracted from \textit{Inventory of Existing Chemical Substances in China}, released by Ministry of Ecology and Environment, China \footnote{\url{https://www.mee.gov.cn/gkml/hbb/bgg/201301/t20130131_245810.htm}}.

For Chinese idioms, we use the crawled data from the Github repo \footnote{\url{https://github.com/pwxcoo/chinese-xinhua}}, and have manually checked the data quality (Of the randomly selected 50 examples, there are only 2 examples that have quality issues).

\begin{figure}
    \centering
    \includegraphics[width=0.95\linewidth]{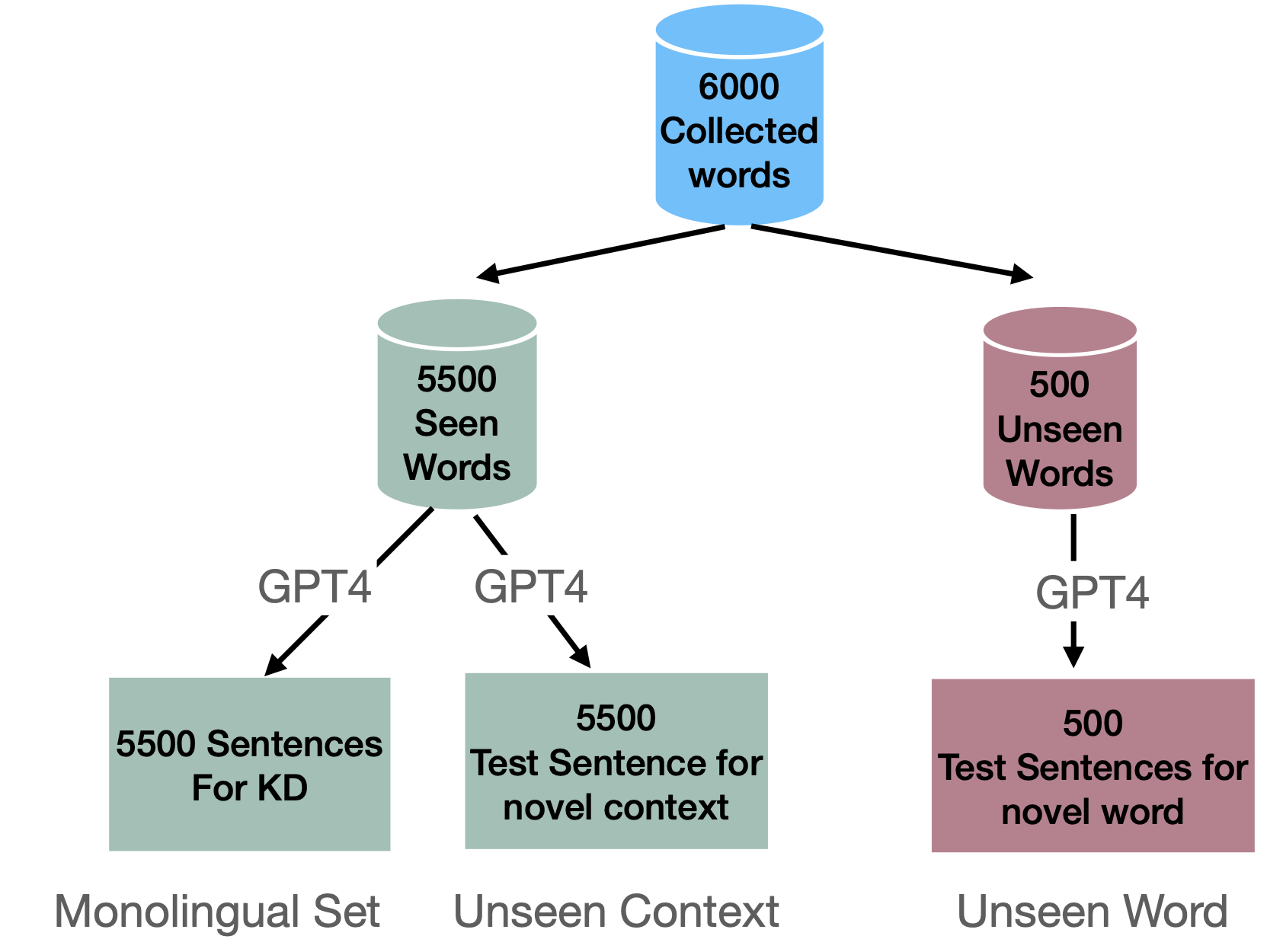}
    \caption{Illustration of the process how the monolingual set and two test sets are splitted from initial collected word sets.}
    \label{fig:testset_flowchart}
\end{figure}

We split each word set to two subsets with 5500 and 500 words, respectively, and use GPT-4 to synthesize contexts for them. Figure~\ref{fig:testset_flowchart} illustrates the process of constructing the monolingual set and two test sets.

\section{Prompts for Evaluation}
Table~\ref{prompt:idiom_evaluation} shows the prompt we used for evaluating the translation quality of Chinese idioms. Table~\ref{prompt:translation_comparison} shows the prompt we used for translation comparison between direct generation and feedback.

\begin{figure*}[]
\small
\begin{mdframed}
\centering
\parbox{\textwidth}{%
        \texttt{
        {Assuming you are a highly proficient translator skilled at providing detailed and comprehensive assessments of machine translations. I will give you a <srclang> sentence X and its <tgtlang> translation Y, and I would like you to help assess the translation. \\
1. You should first provide an overall assessment. \\
2. Following that, \\
    - If there are no errors, just say "No error." and do not provide an explanation. \\
    - If there are errors, please specify  \\
        - the error type, \\
        - the corresponding segment in the <srclang> sentence X, \\
        - the corresponding segment in the translation Y, \\
        - the reason for the error, \\
        - and the correct translation for the segment \\
    - If there are errors, you should also provide a good translation at the end of the assessment. \\
4. For multiple errors, you should address them separately. \\
5. Try to pinpoint the smallest segments containing errors and explain them, avoiding cases where the error encompasses the entire sentence. \\
6. Carefully read the original text and the translation to identify all translation errors. \\
7. Your response should be in English. \\
8. Be concise. \\
\\
Now, please assess the following translation: \\
\\
<srclang>: <srctext> \\
<tgtlang>: <tgttext> \\
\\
Assessment: \\
}}}%
\end{mdframed}
\captionof{table}{Prompt that we use for the feedbacker task.}
\label{prompt:feedback}
\end{figure*}

\begin{figure*}[]
\small
\begin{mdframed}
\centering
\parbox{\textwidth}{%
        \texttt{
        {
        Suppose you are a language expert of <srclang> and <tgtlang>. Given a sentence X, please point out its topic, domain and style. \\
Input: \\
X: <srctext> \\
Output:
}}}%
\end{mdframed}
\captionof{table}{Prompt that we use for the sentence analysis task.}
\label{prompt:sentence_analysis}
\end{figure*}

\begin{figure*}[]
\small
\begin{mdframed}
\centering
\parbox{\textwidth}{%
        \texttt{
        {
Suppose you are a language expert of <srclang> and <tgtlang>. Given a topic, a domain and a style, as well as a bilingual word pair, please generate a pair of parallel sentences that adhere to the given topic, domain and style. They should also contain the given word pair. \\
Input: \\
Domain: <domain> \\
Topic: <topic> \\
Style: <style> \\
Word Pair: <wordpair> \\
\\
Output:
}}}%
\end{mdframed}
\captionof{table}{Prompt that we use for the parallel data synthesizer task.}
\label{prompt:pds}

\end{figure*}

\begin{figure*}[]
\small
\begin{mdframed}
\centering
\parbox{\textwidth}{%
        \texttt{
        {
Assume you are a <srclang> and <tgtlang> language expert with a wealth of knowledge and associative ability in both languages. I will give you a word/phrase P from an <srclang> sentence X. Please associate from the following aspects and generate three words similar to X for each aspect, and provide the <tgtlang> translation of these words. \\
\\
Aspects of association: \\
- Category. What kind of category does this word belong to? \\
- Semantics. What words often appear in the same context as the given word? \\
\\
NOTE, the associated words should be rare words, so that it is unlike for a machine translation system to translate it correctly. \\
\\
Input: \\
X: <srctext> \\
P: <errorword> \\
\\
Output:
}}}%
\end{mdframed}
\captionof{table}{Prompt that we use for the word analogy task.}
\label{prompt:wa}
\end{figure*}

\begin{figure*}[]
\small
\begin{mdframed}
\centering
\parbox{\textwidth}{%
        \texttt{
        {
Assume you are a language expert in English and Chinese. I will give you a Chinese idiom S, a sentence X that contains S, and a machine-generated English translation Y of the source sentence X.
I will also give you the explanation/definition E of the idiom S. Your task is to first identify the translation of S in Y, and judge whether the translation of the idiom is correct. \\
\\
Note:\\
1. The score range is 0/1/2/3/4/5, where \\
- 0: Completely incorrect translation or no translation \\
    - 1: Literal translation of the original, without conveying any implied  meaning, leaving non-Chinese background readers baffled \\
    - 2: Literal translation of the original, partially conveying the implied meaning, easy for non-Chinese background readers to understand \\
    - 3: Interpretative translation of the idiom, but only partially conveying the implied meaning \\
    - 4: Interpretative translation of the idiom, fully conveying the implied meaning \\
    - 5: The translation perfectly conveys the implied meaning of the idiom, is very easy for all readers to understand, and also maintains the aesthetic sense of the original \\
\\
2. You should generate the explanation of your decision concisely. \\
Now, please process the following inputs:\\
}}}%
\end{mdframed}
\captionof{table}{Prompt that we use for evaluating the quality of translating Chinese idioms.}
\label{prompt:idiom_evaluation}
\end{figure*}

\begin{figure*}[]
\small
\begin{mdframed}
\centering
\parbox{\textwidth}{%
        \texttt{
        {
Assume you are a language expert in Chinese and English. 
I will give you a sentence X, the word P in that sentence, and two translations of the sentence X: A and B. Your task is to assess which translation contains the correct translation of the word P. \\
\\
Requirements: \\
(1) Ignore other differences between the two translations. Only compare the translation of the word P. \\
(2) Your answer should first state the reason for your comparison, and then give your comparison. \\
(3) Your comparison should be A, B, C and D. \\
    - A: the first translation of the word P is better. \\
    - B: the second translation of the word P is better. \\
    - C: Both are fine. \\
    - D: Both are bad. \\
\\
Now, please process the following inputs:
}}}%
\end{mdframed}
\captionof{table}{Prompt that we use for comparing translations from direction generation and feedback.}
\label{prompt:translation_comparison}
\end{figure*}

\end{document}